\documentclass[11pt,letterpaper]{article}
\usepackage{emnlp2017}
\usepackage{times}
\usepackage{latexsym}

\usepackage{url}
\usepackage{scrextend} 
\usepackage{color,floatrow}
\usepackage{booktabs}
\usepackage{lingmacros}
\usepackage{xspace}
\usepackage{graphicx}
\usepackage{adjustbox}
\usepackage{bbding} 
\usepackage{subcaption}

\emnlpfinalcopy

\def\emnlppaperid{***}

\title{Is writing style predictive of scientific fraud?}

\author{Chlo{\'e} Braud \and Anders S{\o}gaard \\
        CoAStaL DIKU\\University of Copenhagen\\ University Park 5, 2100 Copenhagen\\ {\tt chloe.braud@gmail.com soegaard@di.ku.dk}}

\date{}

\begin{document}

\maketitle

\begin{abstract}
The problem of detecting scientific fraud using machine learning was recently introduced, with initial, positive results from a model taking into account various general indicators.
The results seem to suggest that writing style is predictive of scientific fraud. 
We revisit these initial experiments, and show that the leave-one-out testing procedure they used likely leads to a slight over-estimate of the predictability, 
but also that simple models can outperform their proposed model by some margin. 
We go on to explore more abstract linguistic features, such as linguistic complexity and discourse structure, only to obtain negative results. 
Upon analyzing our models, we do see some interesting patterns, though: Scientific fraud, for examples, contains less comparison, as well as different types of hedging and ways of presenting logical reasoning.

\end{abstract}

\section{Introduction}

Cases of scientific misconduct are identified every year. 
Scientific papers are retracted because of errors, or for suspected fraud, ranging from plagiarism and minor manipulations to faking the data and disguising the results. 
It has been shown that, however, among the retracted articles indexed in PubMed, only $21.3$\% are retracted due to error, while $67.4$\% were removed due to misconduct, among which suspected fraud amounts to $43.4$\%, the others being due to duplicate publications or plagiarism \cite{fang:misconduct:12}.

In a recent paper, \newcite{markowitz_linguistic_2015} proposed the first analysis of writing style in fraudulent papers across authors and disciplines.
They approached the question of whether these authors have a specific writing style, from a psychological perspective. 
They found that these papers exhibit a higher rate of jargon, make a higher use of references, and have a lower readability rate, suggesting that the authors try to obfuscate their writing, making them harder to read and analyze.
They report classification results using a leave-one-out strategy over the dataset, with a classification accuracy of $57.2$\%. 
As suggested in the paper, we propose to improve this performance by evaluating different classification models.

In this paper, we first show that much better results can be obtained using a simple bag-of-words representation and Logistic Regression. 
Our best model is a syntax-enhanced trigram-model. We also show that the leave-one-out strategy used by the authors leads to an over-estimation of model precision, and we report new results based on a more robust strategy, taking into account the low number of instance available; namely a {\em nested} cross-validation \cite{varma_bias_2006,scheffer_error_1999}. We also considered semantic and discourse features, but we did not observe improvements with such features. 

Of course, that a bag-of-words model outperforms a model based on psychologically motivated features, may simply be the result of overfitting. We present an extensive feature analysis to validate our models, as well as to test psychologically motivated hypotheses from the literature. 

\paragraph{Contributions}
(i) We present a simple model with high accuracy, and show that it implicitly captures the previously-proposed psychologically-motivated features. 
    (ii) We show that adding semantics and discourse features does not lead to improvements.
    (iii) On the other hand, our feature analysis suggests that the models {\em do} learn to focus on concepts that are intuitively related to scientific misconduct, e.g., that scientific fraud contains less comparison.

\section{Related work}

\newcite{markowitz_linguistic_2015} were the first to study writing style in fraudulent papers.
They gathered a corpus of $253$ articles indexed in PubMed that have been retracted for fraudulent data, as well as $253$ unretracted papers (see Section~\ref{sec:data}).
They define five indicators of obfuscation, and show that fraudulent papers tend to demonstrate a higher rate of linguistics obfuscation, corresponding to a lower readability, an higher use of jargon and a higher degree of abstraction.
Linked to studies on deception identification, they also report a lower rate of positive emotion terms and a higher rate of causal terms (e.g. ``depend", ``induce", ``manipulated") in fraudulent papers.
The readability score was computed using Coh-Metrix \cite{mcnamara2013coh}, while the other scores were based on the Linguistic Inquiry and Word Count (LIWC; \cite{pennebaker2007linguistic}), a dictionary associating a word to various scores such as abstraction (a word is considered as jargon if it is not found in the dictionary). 
Finally, they report $57.2$\% in accuracy using these five indicators as features, a score that we show is probably a little too optimistic, since it is based on a leave-one-out procedure (see Section~\ref{sec:expe}).
We extend their work by first showing that a simple unigram model outperforms their model by a large margin, but also by considering more indicators, including discourse and syntax, and by showing, as mentioned, that their scores were probably over-estimated due to their validation strategy.

Our work is also inspired by another related field of research concerned with deception detection. 
\newcite{mihalcea2009lie} 
built three datasets consisting of $100$ true and $100$ deceptive short statements on three different topics (abortion, death penalty, best friend).
Using only unigrams, they report 70.8\% accuracy in a 10-fold cross validation.
They found that specific word classes, as defined in the LIWC, were predictive of deceptive texts, especially classes indicating detachment from self or related to certainty. 

\newcite{feng2012syntactic} investigate syntactic features, using lexicalized and unlexicalized production rules in addition to shallow features (words unigram and bigram, and POS unigram).
They experiment on truthful and deceptive reviews from TripAdvisor, either gold \cite{ott2011finding} or retrieved using a fake review detector \cite{feng2012distributional}, reviews automatically extracted from Yelp, 
and the corpus introduced in \cite{mihalcea2009lie}.
They report scores between $64.3$ and $91.2$\% accuracy, depending on the dataset.
They found that, for all datasets, syntax helps, and 
that deceptive reviews more frequently use VP, SBAR and WHADVP.

We also consider $n$-gram features, syntactic features, as well as discourse features.
Our task is however a bit different, since authors of fraudulent papers are not directly lying, rather trying to conceal their fraud.
Moreover, our documents are longer and are of a different genre, i.e. scientific articles.

\section{Data}
\label{sec:data}

We use the dataset proposed in \cite{markowitz_linguistic_2015} containing $253$ publications retracted for data fraud and $253$ unretracted publications. 
These publications were taken from the PubMed archives from $1973$ through $2013$.

The unretracted papers are extracted by considering one retracted paper and taking a control paper published the same year, in the same journal, and with some common keywords when possible. 
When no such paper exists (around $19$\% of the papers), a paper from an adjacent year, or using the same words in the abstract, was selected.

The data used is the pre-processed version presented in \cite{markowitz_linguistic_2015}: Words were converted from British English to American English forms. Brackets, parentheses, and percent signs were removed. Periods were removed from certains words, such as `Dr.' or Ìnc.'.
The documents only contain the main body text (no section titles, figures, or tables).

\section{Methodology}

We investigate different types of features, from $n$-grams to discourse.
In large vocabulary feature spaces, we perform feature reduction, to reduce sparsity.
We then provide an analysis of the features to identify the most informative indicators.

\paragraph{Word features}

We use word $n$-grams as features, 
with $n \in \{1,2,3\}$. In order to test the hypotheses presented in previous studies, we also use lexicons to extract information about the tokens.
We use the General Inquirer \cite{stone:general:1966} to extract words expressing a \textit{polarity} -- the features built represent the polarity between positive, negative, both and neutral --, and words corresponding to a \textit{causal} term. 
We also use this lexicon to map the words to a more general semantic category (\textit{Inquirer}).

We identify all the personal \textit{pronouns} 
using manually defined lists. 
Finally, we also include as features \textit{hedge} and modal words, also using a pre-defined list.\footnote{\url{https://github.com/wooorm/hedges/blob/master/index.json}}

\paragraph{Syntactic features}

In order to obtain syntactic information, we parse the data using UDPipe\footnote{\url{http://ufal.mff.cuni.cz/udpipe}} \cite{udpipe:2016}, and a prebuilt model available online for English.\footnote{UD 1.2, \url{https://lindat.mff.cuni.cz/repository/xmlui/handle/11234/1-1659}} 
We follow \cite{johannsen:cross:2015} in extracting all subtrees of up to three tokens (\textit{treelets}).

\paragraph{Discourse features}

Finally, we automatically annotate all the data with discourse connectives and explicit discourse relations using simple models trained on the Penn Discourse Treebank (PDTB) \cite{prasad:penn:2008}, a corpus of news articles from the Wall Street Journal.
Discourse coherence is an indicator of the quality of a text  \cite{lin:automatically:2011}, of its reasoning that could reveal an attempt to deceive. Some specific semantic relations could also be good indicators (e.g. Cause).

We used models to identify the discourse connectives (\textit{Connectives}) and to identify the explicit discourse relation\footnote{We ignore the non explicit relations for which the in-domain scores are very low -- around 40-57\% in accuracy \cite{rutherford:improving:2015,lin:pdtb:2014}.} (\textit{Explicit relations}) they trigger, either among the $4$ coarse-grained classes (\textit{lvl1}) at the top of the hierarchy of sense
or using the $11$ more fine-grained relations at the second level  (\textit{lvl2}).
Our models use Logistic Regression and the connective and the surrounding words and their POS as features \cite{lin:recognizing:2009}. 
They are trained on the sections 2-21 of the PDTB.
Our results on the section 23 are close to the state-of-the-art \cite{pitler:using:2009,pitler:easily:2008,lin:pdtb:2014}: 
$92.9$\% in accuracy for identifying the connectives, $95.1$\% for the level-1 relations, and $86.2$\% for the level-2 relations.

\paragraph{Feature analysis}

In addition to presenting accuracies obtained with these feature sets, we also perform a feature analysis. For this purpose we use a combination of correlation coefficients, logistic regression coefficients, and stability selection
\cite{meinshausen2010stability} -- a method that consists in repeatedly fitting the model across different random subsamples, and counting how many times features are selected in $\ell_1$-regularized logistic regression models. 
For stability selection, we use the implementation available in scikit-learn \cite{scikit-learn} with its default parameters, run it on the whole dataset and keep features selected more than $50$\% of the time.

We indicate the size of the original vocabulary and the number of selected features for each category in Table \ref{table:vocab}.

\begin{table}[t!]
\resizebox{\textwidth}{!}{
\begin{tabular}{lrr}
\toprule
Category & \# Orig. feat. & \# Selec. feat. \\
\midrule
Unigrams 		& $65,798$		& $118$	\\
2-3-grams 		& $1,745,188$	& $154$	\\
Polarity 		& $4$		& $-$ \\ 
Causal  		& $68$		& $-$ \\ 
Inquirer 		& $180$ & $-$ \\
Pronouns		& $7$ & $-$ \\
Hedges  		& $121$ 	& $-$ \\
\midrule
Treelets 		& $50,522$	& 	$136$ \\
\midrule
Connectives 			& $70$		& $-$ \\ 
Explicit relations lvl1 	& $4$		& $-$ \\ 
Explicit relations lvl2 	& $10$		& $-$ \\ 
\bottomrule
\end{tabular}}%
\caption{Size of the original vocabulary and number of selected features for $n$-grams and treelets.}
\label{table:vocab}
\end{table}

\section{Classification}
\label{sec:expe}

\paragraph{Representation}

We test separately count vectorizations with each set of features -- unigrams, 2-3-grams, polarity, causality, Inquirer categories, pronouns (grouping per person, or considering each lemma), treelets, connectives, hedge words, level-1 relations and level-2 relations, and combinations of these features.

\paragraph{Model}

We use a binary logistic regression classifier, optimizing the norm ($\ell_1$ or $\ell_2$) and strength ($c \in \{0.001, 0.005, 0.01, 0.1, 0.5, 1, 5, 10, 100\}$ of the regularization term on held-out data.

\begin{figure*}[!ht]
\begin{subfigure}{.3\textwidth}
 \centering
 \includegraphics[width=\linewidth]{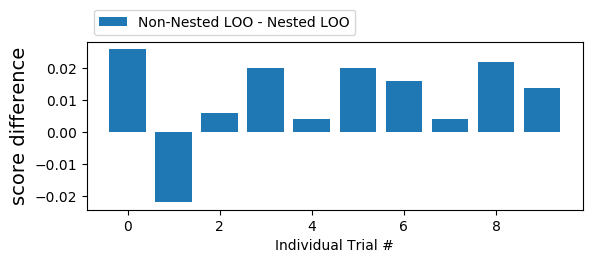}
 \caption{Unigrams}
 \label{fig:sfig1}
\end{subfigure}%
\begin{subfigure}{.3\textwidth}
 \centering
 \includegraphics[width=\linewidth]{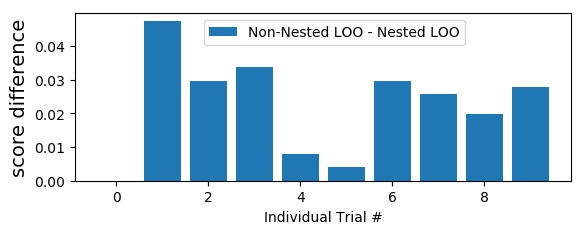}
 \caption{2-3-grams}
 \label{fig:sfig2}
\end{subfigure}
\begin{subfigure}{.3\textwidth}
 \centering
 \includegraphics[width=\linewidth]{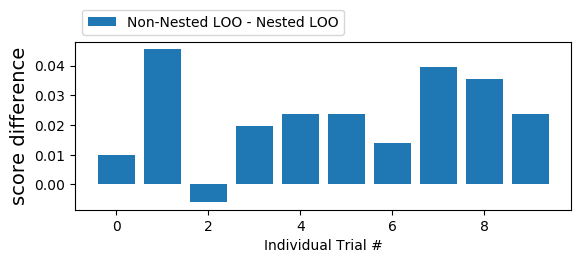}
 \caption{Treelets}
 \label{fig:sfig3}
\end{subfigure}%
\caption{Accuracy difference between LOO and Nested LOO for each trial for different features.}
\label{fig:diffloo}
\end{figure*}

\paragraph{Validation schemes}

\newcite{markowitz_linguistic_2015} report results with a leave-one-out strategy (LOO). 
However, LOO often under-estimates the error rate. 
We compare with a {\em nested}~cross-validation procedure that can provide an almost unbiased estimate of the true error \cite{varma_bias_2006,scheffer_error_1999}.

Specifically, we use two cross-validation loops: the inner loop is used for tuning the hyper-parameters, and the outer loop estimates the generalization error. 
The data are first split into $N$ folds, 
the fold $k$ ($1 \leq k \leq N$ ) is the current evaluation set, and the $N-1$ other folds are used as training data and split into $M$ folds used for model fitting. 
The best model is then evaluated on fold $k$. 
Final scores are averages over the $N$ folds.

For comparison with \newcite{markowitz_linguistic_2015}, we report performance with LOO and with nested cross-validation using 
LOO as outer loop, the inner loop being a random $5$-fold cross-validation.
We repeat each evaluation $10$ times, and report a mean over these trials.

\begin{table}[t!]
\resizebox{\textwidth}{!}{
 	\begin{tabular}{lll}
	\toprule
    System 							& LOO		& N-LOO	 \\ 
	\midrule
	\cite{markowitz_linguistic_2015}& $57.2$ 	& -	 \\ 
  \midrule
  	Unigrams 						& $72.1$ & $71.7$	 \\ 
    2-3-grams 						& $70.8$ & $69.6$    \\ 
    Polarity 					& $50.0$ & $45.3$    \\ 
    Causal 						& $59.9$ & $58.4$    \\
    Inquirer 					& $58.7$ & $54.3$  	 \\ 
    Pronouns 	 				& $54.5$ & $52.2$  	 \\
    Hedges 						& $56.7$ & $54.1$  	 \\
    \midrule
    Treelets 						& $72.9$ & $71.7$	 \\
    \midrule
    Connectives 					& $60.1$ & $58.3$	 \\ 
    Explicit Relations lvl1 		& $54.3$ & $53.2$	 \\ 
    Explicit Relations lvl2 		& $54.5$ & $54.3$  	 \\
    \midrule
    \midrule
	1-2-3-grams+treelets 			& $\mathbf{76.3}$ & $\mathbf{76.0}$ 	 \\ 
    All 							& $70.3$ & $69.8$ \\
	\bottomrule
	\end{tabular}}%
    \caption{Results (accuracy, in \%).}%
    \label{tab:results}
\end{table}

\paragraph{Results}

Our results are summarized in Table \ref{tab:results}.
Our results are generally higher than the $57.2$\% reported in \cite{markowitz_linguistic_2015}, with at best $71.7$\% with a nested LOO and a single group of features (\textit{unigrams} or \textit{treelets}) and $76.0$\% when $n$-grams and treelets are combined.

Using all the $n$-grams already leads to a better accuracy score ($+1.3$\%) compared to using only \textit{unigrams} ($73.0$\% in accuracy for \textit{1+2-3-grams} with N-LOO).
On the other hand, combining discourse features to the $n$-grams does not allow improvements over using only the $n$-grams ($72.8$\% with N-LOO for \textit{1+2-3-grams+Connectives+Explicit Relations lvl1}).

The scores obtained with LOO are over-estimate performance, compared to  nested cross-validation, see for example Figure \ref{fig:diffloo}: Even if the differences are low, they are consistent across the trials and the feature sets.

\section{Feature analysis}

We use Pearson's $\rho$ (w.~Bonferroni correction) to establish what features are predictive of fraud and non-fraud.
We report the values for the features cited in Table~\ref{tab:indicators}.

\paragraph{Hedging} There is an interesting contrast between adverbial hedges ({\em conceivably, presumably, surely, effectively}) and verbal hedges ({\em suggest}) indicative of fraud, and adverbial hedges ({\em practically, occasionally}) and verbal hedges {\em assume, speculate}) indicative of non-fraud: It seems adverbs and verbs used in fraud are for interpreting the data on behalf of the reader, whereas the adverbs and verbs indicative of fraud are more observer-aware (e.g.,~{\em we speculate}). This suggest that a fraud strategy is to hide observer’s bias, rather than being explicit about it.

\paragraph{Comparison} Both the discourse relation and the Inquirer class for comparison are predictive of non-fraud. Scientific fraud thus seems less likely to compare.
On the other hand, neither  the causal relations or the presence of causal terms were significantly linked to fraudulent papers.

\paragraph{Therefore vs. since} A peculiar, but statistically significant difference between fraud and non-fraud articles, is that fraud articles prefer {\em therefore} over {\em since}, and vice versa. We speculate that it may be a fraud strategy to make the reasoning more verbose by separating out premises  (because the authors are, consciously or not, afraid the readers will not accept them). This is in slight contrast with or qualifies the main hypothesis in \newcite{markowitz_linguistic_2015}, that fraudulent writers try to obfuscate their writing.

\paragraph{Other markers of fraud} Many technical concepts were highly correlated with fraud, but we suspect these are cases of overfitting. More interestingly, the bigram {\em described previously} was among the top-5 most highly correlated features, indicating fraud. From our syntactic treelets, proper nouns and interjections were both slightly indicative of fraud ($p<0.01$). 

\paragraph{Other markers of non-fraud} From our syntactic treelets, conjunctions of numbers were indicative of non-fraud, suggesting maybe a higher level of technical detail. Non-fraud articles are also more likely to use the pronoun {\em they}, as compared to {\em we}, compared to fraud papers.

\begin{table}[t!]
\resizebox{\textwidth}{!}{
 	\begin{tabular}{llll}
 	\toprule
\multicolumn{3}{c}{Hedges} \\
	assume & -0.121 & p=0.006 \\
	practically & -0.118 & p=0.008 \\
	occasionally & -0.112 & p=0.012 \\
	conceivably & 0.089 & p=0.045 \\
	assumed & -0.086 & p=0.052 \\
	surely & 0.077 & p=0.083 \\
	effectively & 0.075 & p=0.090 \\
	presumably & 0.058 & p=0.195 \\
\midrule
\multicolumn{3}{c}{Inquirer} \\
	compare & -0.158 & p=0.0003 \\
\midrule
\multicolumn{3}{c}{Explicit Relations lvl1} \\
	comparison & -0.096 & p=0.031 \\
    cause & 0.008 & p=0.863 \\
    \midrule
    \multicolumn{3}{c}{Connectives} \\
    since & -0.102 & p=0.022 \\
    therefore & 0.064 & p=0.147 \\
    \midrule
    \multicolumn{3}{c}{2-3-grams} \\
    described previously & 0.115 & p=0.009 \\
    \midrule
    \multicolumn{3}{c}{Treelets} \\
    intj & 0.126 & p=0.004 \\
    propn & 0.110 & p=0.013 \\
    \midrule
    \multicolumn{3}{c}{Pronouns} \\
    we & 0.071 & p=0.112 \\
	they & -0.059 & p=0.182 \\
	\end{tabular}}%
     \caption{Pearson $\rho$ and original $p$-value (before Bonferroni correction) for some features.}
     \label{tab:indicators}
 \end{table}

\section{Conclusion}

We show that a simple unigram model outperforms previous work on scientific fraud detection. Overall, more high-level linguistic features, beyond syntactic treelets, do not lead to improvements, but we also presented a feature analysis showing, for example, that comparison and explanation (at the semantic and discourse level) are indicators of non-fraud, and that fraudulent writing uses slightly different hedging strategies.

\bibliography{biblio}

\begin{thebibliography}{22}
\expandafter\ifx\csname natexlab\endcsname\relax\def\natexlab#1{#1}\fi

\bibitem[{Fang et~al.(2012)Fang, Steen, and Casadevall}]{fang:misconduct:12}
Ferric~C. Fang, R.~Grant Steen, and Arturo Casadevall. 2012.
\newblock {Misconduct accounts for the majority of retracted scientific
  publications.}
\newblock \emph{Proceedings of the National Academy of Sciences of the United
  States of America}, 109(42):17028--17033.

\bibitem[{Feng et~al.(2012{\natexlab{a}})Feng, Banerjee, and
  Choi}]{feng2012syntactic}
Song Feng, Ritwik Banerjee, and Yejin Choi. 2012{\natexlab{a}}.
\newblock Syntactic stylometry for deception detection.
\newblock In \emph{Proceedings of the 50th Annual Meeting of the Association
  for Computational Linguistics: Short Papers-Volume 2}, pages 171--175.
  Association for Computational Linguistics.

\bibitem[{Feng et~al.(2012{\natexlab{b}})Feng, Xing, Gogar, and
  Choi}]{feng2012distributional}
Song Feng, Longfei Xing, Anupam Gogar, and Yejin Choi. 2012{\natexlab{b}}.
\newblock Distributional footprints of deceptive product reviews.
\newblock \emph{ICWSM}, 12:98--105.

\bibitem[{Johannsen et~al.(2015)Johannsen, Hovy, and
  S{\o}gaard}]{johannsen:cross:2015}
{Anders Trærup} Johannsen, Dirk Hovy, and Anders S{\o}gaard. 2015.
\newblock Cross-lingual syntactic variation over age and gender.
\newblock In \emph{Proceedings of the Nineteenth Conference on Computational
  Natural Language Learning}.

\bibitem[{Lin et~al.(2009)Lin, Kan, and Ng}]{lin:recognizing:2009}
Ziheng Lin, Min-Yen Kan, and Hwee~Tou Ng. 2009.
\newblock Recognizing implicit discourse relations in the penn discourse
  treebank.
\newblock In \emph{Proceedings of EMNLP}.

\bibitem[{Lin et~al.(2011)Lin, Ng, and Kan}]{lin:automatically:2011}
Ziheng Lin, Hwee~Tou Ng, and Min-Yen Kan. 2011.
\newblock Automatically evaluating text coherence using discourse relations.
\newblock In \emph{Proceedings of ACL-HLT}.

\bibitem[{Lin et~al.(2014)Lin, Ng, and Kan}]{lin:pdtb:2014}
Ziheng Lin, Hwee~Tou Ng, and Min-Yen Kan. 2014.
\newblock A pdtb-styled end-to-end discourse parser.
\newblock \emph{Natural Language Engineering}, 20:151--184.

\bibitem[{Markowitz and Hancock(2015)}]{markowitz_linguistic_2015}
David~M. Markowitz and Jeffrey~T. Hancock. 2015.
\newblock {Linguistic Obfuscation in Fraudulent Science}.
\newblock \emph{Journal of Language and Social Psychology}.

\bibitem[{McNamara et~al.(2013)McNamara, Louwerse, Cai, and
  Graesser}]{mcnamara2013coh}
DS~McNamara, MM~Louwerse, Z~Cai, and A~Graesser. 2013.
\newblock Coh-metrix version 3.0.
\newblock \emph{Retrieved [4/1/15] from http://cohmetrix. com}.

\bibitem[{Meinshausen and B{\"u}hlmann(2010)}]{meinshausen2010stability}
Nicolai Meinshausen and Peter B{\"u}hlmann. 2010.
\newblock Stability selection.
\newblock \emph{Journal of the Royal Statistical Society: Series B (Statistical
  Methodology)}, 72(4):417--473.

\bibitem[{Mihalcea and Strapparava(2009)}]{mihalcea2009lie}
Rada Mihalcea and Carlo Strapparava. 2009.
\newblock The lie detector: Explorations in the automatic recognition of
  deceptive language.
\newblock In \emph{Proceedings of the ACL-IJCNLP 2009 Conference Short Papers},
  pages 309--312. Association for Computational Linguistics.

\bibitem[{Ott et~al.(2011)Ott, Choi, Cardie, and Hancock}]{ott2011finding}
Myle Ott, Yejin Choi, Claire Cardie, and Jeffrey~T Hancock. 2011.
\newblock Finding deceptive opinion spam by any stretch of the imagination.
\newblock In \emph{Proceedings of ACL HLT}.

\bibitem[{Pedregosa et~al.(2011)Pedregosa, Varoquaux, Gramfort, Michel,
  Thirion, Grisel, Blondel, Prettenhofer, Weiss, Dubourg, Vanderplas, Passos,
  Cournapeau, Brucher, Perrot, and Duchesnay}]{scikit-learn}
F.~Pedregosa, G.~Varoquaux, A.~Gramfort, V.~Michel, B.~Thirion, O.~Grisel,
  M.~Blondel, P.~Prettenhofer, R.~Weiss, V.~Dubourg, J.~Vanderplas, A.~Passos,
  D.~Cournapeau, M.~Brucher, M.~Perrot, and E.~Duchesnay. 2011.
\newblock Scikit-learn: Machine learning in {P}ython.
\newblock \emph{Journal of Machine Learning Research}, 12:2825--2830.

\bibitem[{Pennebaker et~al.(2007)Pennebaker, Booth, and
  Francis}]{pennebaker2007linguistic}
James~W Pennebaker, Roger~J Booth, and Martha~E Francis. 2007.
\newblock Linguistic inquiry and word count: Liwc.
\newblock \emph{Austin, TX: Pennebaker Conglomerates}.

\bibitem[{Pitler and Nenkova(2009)}]{pitler:using:2009}
Emily Pitler and Ani Nenkova. 2009.
\newblock Using syntax to disambiguate explicit discourse connectives in text.
\newblock In \emph{Proceedings of the ACL-IJCNLP}.

\bibitem[{Pitler et~al.(2008)Pitler, Raghupathy, Mehta, Nenkova, Lee, and
  Joshi}]{pitler:easily:2008}
Emily Pitler, Mridhula Raghupathy, Hena Mehta, Ani Nenkova, Alan Lee, and
  Aravind Joshi. 2008.
\newblock Easily identifiable discourse relations.
\newblock In \emph{Proceedings of COLING (Posters)}.

\bibitem[{Prasad et~al.(2008)Prasad, Dinesh, Lee, Miltsakaki, Robaldo, Joshi,
  and Webber}]{prasad:penn:2008}
Rashmi Prasad, Nikhil Dinesh, Alan Lee, Eleni Miltsakaki, Livio Robaldo,
  Aravind Joshi, and Bonnie Webber. 2008.
\newblock The {P}enn {D}iscourse {T}reebank 2.0.
\newblock In \emph{Proceedings of LREC}.

\bibitem[{Rutherford and Xue(2015)}]{rutherford:improving:2015}
Attapol Rutherford and Nianwen Xue. 2015.
\newblock Improving the inference of implicit discourse relations via
  classifying explicit discourse connectives.
\newblock In \emph{Proceedings of NAACL-HLT}.

\bibitem[{Scheffer(1999)}]{scheffer_error_1999}
Tobias Scheffer. 1999.
\newblock \emph{Error Estimation and Model Selection}.
\newblock Ph.D. thesis, Technischen Universitet Berlin, School of Computer
  Science.

\bibitem[{Stone and Kirsh(1966)}]{stone:general:1966}
Philip~J. Stone and John Kirsh. 1966.
\newblock \emph{The General Inquirer: A Computer Approach to Content Analysis}.
\newblock MIT Press.

\bibitem[{Straka et~al.(2016)Straka, Haji\v{c}, and Strakov\'{a}}]{udpipe:2016}
Milan Straka, Jan Haji\v{c}, and Strakov\'{a}. 2016.
\newblock {UDPipe:} {T}rainable {P}ipeline for {P}rocessing {CoNLL-U} {F}iles
  {P}erforming {T}okenization, {M}orphological {A}nalysis, {POS} {T}agging and
  {P}arsing.
\newblock In \emph{Proceedings of the Tenth International Conference on
  Language Resources and Evaluation (LREC'16)}.

\bibitem[{Varma and Simon(2006)}]{varma_bias_2006}
Sudhir Varma and Richard Simon. 2006.
\newblock Bias in error estimation when using cross-validation for model
  selection.
\newblock \emph{BMC bioinformatics}.

\end{thebibliography}
\bibliographystyle{emnlp_natbib}

\end{document}